MACHINE LEARNING BASED ENGLISH SENTIMENT ANALYSIS

Tran Thi Ngoc Thao[1], Nguyen Ngoc Kim Lien[1], Ngo Minh Vuong[2,*]

*[1]Faculty of Information Technology, Ton Duc Thang University*
*19 Nguyen Huu Tho Street, District 7, Ho Chi Minh City*

*[2]Faculty of Information Technology, Lac Hong University*
*10 Huynh Van Nghe Street, Bien Hoa City, Dong Nai Province*

[*]Email*: ngovuong@lhu.edu.vn*

Sentiment analysis or opinion mining aims to determine attitudes, judgments and opinions of customers for a product or a service. This is a great system to help manufacturers or servicers know the satisfaction level of customers about their products or services. From that, they can have appropriate adjustments. We use a popular machine learning method, being Support Vector Machine, combine with the library in Waikato Environment for Knowledge Analysis (WEKA) to build Java web program which analyzes the sentiment of English comments belongs one in four types of woman products. That are dresses, handbags, shoes and rings. We have developed and test our system with a training set having 300 comments and a test set having 400 comments. The experimental results of the system about precision, recall and F measures for positive comments are 89.3%, 95.0% and 92,.1%; for negative comments are 97.1.%, 78.5% and 86.8%; and for neutral comments are 76.7%, 86.2% and 81.2%.

*Keywords:* sentiment analysis, support vector machine, opinion mining, text classification.

# PHÂN TÍCH Ý KIẾN CỦA NHẬN XÉT TIẾNG ANH DỰA TRÊN PHƯƠNG PHÁP HỌC MÁY

**Trần Thị Ngọc Thảo[1], Nguyễn Ngọc Kim Liên[1], Ngô Minh Vương[2,\*]**

*[1]Khoa Công nghệ thông tin, Trường Đại học Tôn Đức Thắng
19 Nguyễn Hữu Thọ, Quận 7, TpHCM*

*[2]Khoa Công nghệ thông tin, Trường Đại học Lạc Hồng
10 Huỳnh Văn Nghệ, Biên Hòa, Đồng Nai*

*\*Email: ngovuong@lhu.edu.vn*

## TÓM TẮT

Phân tích ý kiến là nhằm mục đích xác định thái độ, quan điểm của người sử dụng đối với một sản phẩm hoặc một dịch vụ nào đó. Đây là một công trình giúp ích cho những nhà sản xuất hoặc nhà cung cấp dịch vụ biết được mức độ hài lòng của khách hàng về sản phẩm hoặc dịch vụ của họ để từ đó có mức độ điều chỉnh thích hợp. Chúng tôi sử dụng phương pháp phổ biến của học máy là máy vectơ hỗ trợ kết hợp với thư viện phần mềm WEKA (Waikato Environment for Knowledge Analysis) để xây dựng chương trình trên nền web Java phân tích ý kiến của nhận xét tiếng Anh về một trong bốn loại sản phẩm dành cho nữ giới là đầm, túi xách, giày và nhẫn. Chúng tôi đã xây dựng và kiểm thử chương trình với tập huấn luyện gồm 300 nhận xét, tập kiểm tra gồm 400 nhận xét. Kết quả đánh giá hiệu suất của hệ thống về độ chính xác, độ đầy đủ và độ F của nhận xét tích cực là 89,3%, 95,0% và 92,1%; của nhận xét tiêu cực là 97,1.%, 78,5% và 86,8%; và của nhận xét trung tính là 76,7%, 86,2% và 81,2%.

*Từ khóa:* Phân tích nhận xét, máy vectơ hỗ trợ, khai phá ý kiến, phân loại văn bản.

## 1. GIỚI THIỆU

### 1.1. Khái niệm ý kiến

Ý kiến của nhận xét, bình luận hay đánh giá của một người về một sản phẩm hoặc dịch vụ là những suy nghĩ, nhận định, tình cảm tâm lý hoặc cảm xúc của người đó về chất lượng, hình dáng, vật liệu hoặc giá cả của sản phẩm hoặc dịch vụ đó. Các ý kiến có thể là tích cực, tiêu cực hoặc trung lập tùy thuộc vào cảm nhận của từng cá nhân.

Thông thường, một đánh giá sẽ gồm các phần như sau:

- Tiêu đề: mô tả ngắn gọn về đánh giá hoặc tên sản phẩm.

- Loại sản phẩm: sản phẩm thuộc về loại nào, ví dụ như đầm hoặc túi xách.

- Sản phẩm: sản phẩm sẽ được đánh giá.

- Nội dung: ý kiến chi tiết của người dùng về sản phẩm.
- Người đánh giá: người viết ý kiến đánh giá.
- Mức đánh giá: thang điểm người đánh giá dành cho sản phẩm dựa trên mức độ hài lòng (1 sao, 2 sao, 3 sao, 4 sao hoặc 5 sao).
- Thời gian đánh giá: ngày tháng năm mà ý kiến đánh giá được viết.

Ngoài ra còn có thêm phần viết phản hồi cho đánh giá đó, đánh giá đó có hữu ích hay không, số người cảm thấy đánh giá đó hữu ích. Với mỗi ý kiến đánh giá, công trình chỉ quan tâm đến loại sản phẩm, tiêu đề và nội dung ý kiến đánh giá.

## 1.2. Các loại ý kiến

Tương tự như công trình [1], chúng tôi phân loại ý kiến thành 3 loại là ý kiến tích cực (positive), ý kiến tiêu cực (negative) và ý kiến trung lập (neutral). Ý kiến tích cực (positive) là ý kiến chứa các thông tin tích cực về sản phẩm, những phản hồi đánh giá tốt, khen ngợi sản phẩm đó. Sản phẩm càng có nhiều ý kiến đánh giá tích cực thì người đọc sẽ càng cảm thấy sản phẩm đó tốt và chất lượng. Trong ý kiến tích cực thường xuất hiện các tính từ chỉ tính chất tính cực của sản phẩm như "beautiful" (đẹp), "pretty" (xinh), "cute" (dễ thương), "good" (tốt), "great" (tuyệt), "nice" (tốt), "well" (tốt), "comfortable" (thoải mái) và các từ chỉ cảm xúc tích cực của người viết như "love" (yêu), "like" (thích), "happy" (hạnh phúc), "glad" (vui), "pleased" (hài lòng), "excited" (phấn khởi), "expect" (mong đợi), "recommend" (đề nghị).

Ý kiến tiêu cực (negative) là ý kiến chứa các thông tin tiêu cực về sản phẩm, những phản hồi đánh giá không tốt về sản phẩm đó. Sản phẩm có chất lượng thấp hoặc kiểu dáng chưa đẹp thường nhận được những ý kiến tiêu cực bởi vì khách hàng cảm thấy không hài lòng. Vì thế các nhà sản xuất sẽ xem xét, ghi nhận các ý kiến đánh giá để thay đổi, cải tiến các sản phẩm này cho tốt hơn. Trong ý kiến tiêu cực thường xuất hiện các tính từ chỉ tính chất tiêu cực của sản phẩm như "tight" (chật), "stiff" (cứng), "poor" (tệ), "wrong" (sai), "weird" (kỳ dị) và các từ chỉ cảm xúc tiêu cực của người viết như "hate" (ghét), "disappointed" (thất vọng), "stupid" (ngớ ngẩn), "return" (trả lại).

Ý kiến trung lập (neutral) là ý kiến đánh giá chung chung về sản phẩm, có thể chứa cả nhận định tích cực và tiêu cực nhưng ở mức độ cân bằng. Ý kiến trung lập có thể chứa nội dung vừa khen vừa chê một sản phẩm, cũng có thể không khen không chê sản phẩm đó. Các ý kiến chứa các từ "ok" (tạm được), "okay" (tạm được), "alright" (ổn) thường là các ý kiến trung lập. Ý kiến trung lập thì không phân cực.

## 2. CÔNG TRÌNH LIÊN QUAN

### 2.1. Phân tích ý kiến dự đoán bầu cử trên web bằng hệ thống Crystal

Công trình [2] trình bày một hệ thống dự đoán bầu cử (Crystal) dựa trên các ý kiến đăng trên trang web dự đoán bầu cử. Để xây dựng hệ thống này, trước tiên phải thu thập các ý kiến về những bầu cử đã qua trên trang web dự đoán bầu cử và tạo thành một tài liệu dữ liệu gọi là corpus. Sau đó sử dụng dữ liệu này cho hệ thống phân tích ý kiến để dự đoán kết quả bầu cử. Với mỗi tin nhắn dự đoán, Crystal đầu tiên xác nhận đảng mà tin nhắn đó dự đoán chiến thắng và sau đó tập hợp các ý kiến đã phân tích thành kết quả dự đoán bầu cử. Để dự đoán kết quả bầu cử, công trình áp dụng Support Vector Machine (SVM) based supervised learning. Ba bước phân tích dự đoán là: tổng quát tính chất, phân loại bằng cách sử dụng SVM, tích hợp kết quả SVM.

Trong bước tổng quát tính chất, hệ thống sẽ khái quát mẫu của các từ được sử dụng trong các ý kiến dự đoán. Vd: Thay vì sử dụng "Đảng Tự do sẽ thắng", "Đảng Dân chủ sẽ thắng", "Đảng Bảo thủ sẽ thắng" thì sẽ khái quát thành "Đảng chiến thắng sẽ thắng". Trước tiên, hệ thống sẽ thay thế cả họ tên ứng viên thành tên của Đảng phái chính trị mà ứng cử viên đó thuộc về. Sau đó hệ thống sẽ ngắt mỗi thông điệp thành nhiều câu. Tiếp theo với mỗi câu hệ thống sẽ lặp lại số lần bằng số Đảng phái xuất hiện trong câu đó, đồng thời thay thế tên các Đảng bằng Đảng chiến thắng và Đảng khác để tổng quát tính chất.

Trong bước phân loại bằng cách sử dụng SVM, sau khi đã có được mẫu (hóa trị, câu tổng quát) trong bước tổng quát tính chất, hệ thống sẽ phân loại câu tổng quát thành hóa trị hướng tới mục tiêu cuối cùng xác định (hóa trị, đảng phái) của thông điệp. Công việc này là một loại phân loại nhị phân vì hóa trị chỉ có 2 loại: +1 và -1. Vd: Với 1 câu tổng quát "Đảng khác như là như sẽ không vượt qua được Đảng chiến thắng", mục tiêu của hệ thống là đặt Đảng chiến thắng hóa trị THẮNG. Hệ thống sẽ thực hiện mô hình SVM bằng việc sử dụng gói SVM light.

Trong bước tích hợp kết quả SVM, hệ thống sẽ kết hợp hóa trị của mỗi câu dự đoán bởi SVM để xác định kết luận hóa trị được dự đoán của thông điệp. Đối với mỗi Đảng có trong thông điệp, hệ thống tính toán tổng các hóa trị từng câu của Đảng đó và chọn Đảng có giá trị cao nhất.

Để cải thiện hiệu suất, công trình áp dụng kỹ thuật mới là tổng quát mô hình tính năng n-gram. Kết quả thực nghiệm cho thấy Crystal nhanh hơn một số đường cơ sở có phương pháp tiếp cận non-generalized n-gram và dự đoán bầu cử chính xác đến 81.68%. Trong tương lai sẽ thực hiện mô hình ý kiến dự đoán trong các lĩnh vực khác như thị trường bất động sản, những lĩnh vực mà yêu cầu nhiều hơn về khai thác, thiết kế hệ thống và thu thập dữ liệu.

## 2.2. Phân tích ý kiến, tình cảm bằng cách sử dụng Appraisal Groups

Công trình [3] trình bày một phương pháp mới để phân loại tình cảm dựa trên khai thác và phân tích ngữ nghĩa chi tiết các thái độ biểu hiện, và đơn vị của các biểu hiện này không phải là các từ riêng lẻ mà là các nhóm thẩm định, ví dụ như very good (rất tốt), hoặc not extremely brilliant (không cực kỳ rực rỡ). Một biểu thức thẩm định có ít nhất các giá trị: Người thẩm định, Vật bị thẩm định, Loại thẩm định, và Định hướng (tích cực/tiêu cực). Công trình này sử dụng một kỹ thuật bán tự động để xây dựng một vốn từ vựng cho các giá trị thuộc tính thẩm định với các điều kiện có liên quan. Phương pháp chuẩn để đặc trưng cho văn bản là vectơ đa chiều có đầu vào là các kỹ thuật học máy để đo tần số của các yếu tố văn bản (như là từ). Phương pháp này được xem là tốt, giúp xác định các tính năng khác nhau của đơn vị từ vựng hoặc các giá trị thuộc tính của nhóm thẩm định.

Hiện nay, thách thức lớn của công trình là xác định chính xác đầy đủ các biểu hiện đánh giá liên quan bao gồm người thẩm định, vật bị thẩm định, loại thẩm định và định hướng. Điều này có thể phân tích mịn hơn trong những biểu hiện tâm lý trong một tài liệu. Lý thuyết ngôn ngữ hiện tại chẳng hạn lý thuyết thẩm định có thể cung cấp cơ sở tính năng mới của văn bản, có thể cải thiện kết quả của các kỹ thuật truyền thống.

## 2.3. Các công trình liên quan khác

Công trình [4] trình bày phương pháp tiếp cận theo hướng tự động phát hiện cảm xúc ẩn, EmotiNet, dựa trên kiến thức thông thường. EmotiNet là một cơ sở kiến thức các khái niệm liên quan đến giá trị tình cảm. Bằng cách sử dụng tài nguyên này, chúng ta có thể phát hiện cảm xúc tiềm ẩn có trong văn bản. EmotiNet được xây dựng bằng các bước sau: thiết kế một ontology chứa các định nghĩa khái niệm chính, mở rộng ontology bằng cách sử dụng các tình huống được lưu trữ trong ISEAR, mở rộng ontology bằng cách sử dụng cơ sở kiến thức thông thường hiện có. Công trình này có thể phát hiện các biểu hiện tâm lý, cảm xúc tiềm ẩn mà không cần có sự xuất hiện của các từ biểu hiện tình cảm, và khai thác phương pháp ontology.

Công trình [5] giải quyết vấn đề này ở cấp độ câu và xây dựng một hệ thống dựa trên quy tắc bằng cách sử dụng Gate framework. Hệ thống phân tích ý kiến về chất lượng sản phẩm này được phân chia thành 2 phần: từ và các cụm từ xác định chứa ý kiến và phân loại câu và tài liệu dựa trên ý kiến. Không giống với phân loại theo loại từ và chủ đề, việc phân loại nhận xét đòi hỏi sự hiểu biết về xu hướng tình cảm trong bài viết. Những khía cạnh thách thức trong việc phân tích ý kiến bao gồm việc xác định ý kiến, cường độ tình cảm, sự phức tạp của câu, từ trong ngữ cảnh khác nhau. Công trình sử dụng phương pháp tách từ, gán nhãn và luật.

Công trình [6] dự vào các từ hoặc cụm từ để phân tích ý kiến theo mức độ năng nhẹ, công trình sử dụng các công cụ trong phân loại ý kiến như Stanford Parser, SentiWordNet, Coreferency Resolution. Trong đó, Standford Parser sử dụng phương pháp cây phân tích. SentiWordNet là kỹ thuật phân loại chi tiết đòi hỏi khả năng tính toán ý kiến dựa trên mỗi từ. Coreference Resolution cho thấy nhiều khía cạnh của ngôn ngữ tự nhiên, Coreference Resolution gần như không hoàn hảo, hiệu xuất làm việc rất hạn chế. Công trình này chủ yếu dùng phương pháp phân tích ngôn ngữ tự nhiên.

Công trình [7] phân tích ý kiến bằng cách trích xuất tình cảm hoặc ý kiến về một chủ đề từ các tài liệu văn bản trực tuyến. Thay vì phân loại tình cảm của toàn bộ chủ đề, phân tích ý kiến phát hiện tất cả các tài liệu tham khảo liên quan đến chủ đề cho trước và xác định tình cảm trong mỗi loại tài liệu tham khảo đó sử dụng kỹ thuật phân tích ngôn ngữ tự nhiên. Công trình này bao gồm: trích xuất các đặc trưng của chủ đề cụ thể; trích xuất tình cảm của mỗi cụm từ biểu hiện tình cảm; liên kết chủ đề hoặc tính chất của chủ đề với ý kiến.

Công trình [8] mặc dù phân tích về các nhận xét tiếng Việt và sử dụng phương pháp học máy. Tuy nhiên, công trình này tìm kiếm các nhận xét rác, tức là các nhận xét không thực, nhận xét quảng cáo và nhận xét nhằm chủ đề chứ không phải phân loại các nhận xét thành nhận xét tích cực, tiêu cực hoặc trung tính.

Công trình [9] sử dụng giải thuật mô hình hóa chủ đề (topic modeling) khai thác ontology để có thể nắm bắt được thông tin tường minh và cũng như tiềm ẩn của sản phẩm. Đồng thời công trình cũng đã phát triển một phương pháp phân tích xã hội để rút trích hiệu quả các bình luận của khách hàng trên mạng xã hội. Cũng khai thác ontology, các công trình [10], [11] và [12] sử dụng thực thể có tên để mở rộng truy vấn hoặc tài liệu nhằm tăng hiệu quả truy hồi thông tin.

## 3. CƠ SỞ LÝ THUYẾT

### 3.1. Học máy

Học máy (machine learning) [13] là một nhánh thuộc trí tuệ nhân tạo, sử dụng kiến thức của xác suất thống kê và đại số tuyến tính để nghiên cứu phát triển các kỹ thuật và xây dựng các chương trình mà máy tính có thể học hỏi, phân tích các dữ liệu đã có và tạo ra kết quả cho các dữ liệu mới. Học máy có rất nhiều ứng dụng thực tế như phân tích ý kiến đánh giá, phân tích thị trường, phát hiện thẻ tính dụng giả, chẩn đoán y khoa, phân loại chuỗi DNA, nhận dạng đối tượng, nhận dạng tiếng nói và chữ viết, phân loại email, dịch tự động, cử động robot.

Một số dạng thuật toán thường dùng được phân loại dựa trên kết quả đầu ra mong muốn hoặc loại đầu vào trong quá trình huấn luyện máy là: học có giám sát (supervised learning), học không giám sát (unsupervised learning), học bán giám sát (semi-supervised), học tăng cường (reinforcement learning), học cách học (learning to learn).

- Học có giám sát là tạo ra một hàm ánh xạ dữ liệu đầu vào tới kết quả mong muốn, tức là xây dựng chương trình xuất ra kết quả cho những dữ liệu đầu vào mới dựa vào những dữ liệu đầu vào đã có đầu ra được gán nhãn từ trước.

  Vd: trong phân loại (classification) và hồi quy (regression).

- Học không giám sát là tập hợp, gom nhóm các dữ liệu đầu vào thành các cụm đầu ra có cùng đặc tính mà không có sẵn các dữ liệu đánh nhãn.

  Vd: trong vấn đề gom cụm (clustering).

- Học bán giám sát là kết hợp cả dữ liệu có gắn nhãn và không gắn nhãn để tạo một hàm hoặc một bộ phân loại thích hợp.

- Học tăng cường là thực hiện hành động tiếp theo dựa vào dữ liệu được ghi nhận từ kết quả của hành động trước đó.

  Vd: trong lập kế hoạch (planning).

- Học cách học là dựa vào những dữ liệu trường hợp đã gặp để dự đoán kết quả cho những trường hợp chưa gặp.

## 3.2. Support Vector Machine

Support Vector Machine (SVM), được giới thiệu bởi Cortes và Vapnik vào năm 1995 [14], đã giúp phát triển thuật toán từ các số liệu thống kê từ thập kỉ 60, được áp dụng trong nhiều lĩnh vực như trong lĩnh vực sử dụng tin học, phân tích dữ liệu sinh học, văn bản và nhận diện hình ảnh.

SVM là phương pháp học được sử dụng cho phân loại nhị phân. Mỗi mẫu dữ liệu này sẽ được biểu diễn thành dạng vector và được ánh xạ thành một điểm trong không gian. Ý tưởng chính là tìm ra mặt phẳng phân chia dữ liệu trong không gian $d$ thành 2 lớp sao cho khoảng cách từ các điểm dữ liệu gần nhất tới mặt phẳng là xa nhất. SVM giới thiệu một khái niệm mới về việc đưa dữ liệu vào trong một không gian có chiều cao hơn nơi mà dữ liệu được phân chia. Hình 1 mô tả việc lựa chọn mặt phẳng có thể làm tăng khoảng cách tối đa của biên độ. Có rất nhiều mặt phẳng có thể phân tách các điểm dữ liệu thành 2 miền nhưng mặt siêu phẳng tối ưu là mặt phẳng có độ lỗ lớn nhất.

Ta có tập huấn luyện D gồm n điểm: D={$X_i, Y_i$}, với i = 1,...,n. $X_i$ thuộc $R_d$ và lớp nhãn $Y_i$ thuộc {-1, 1}. Giả sử những điểm trên D là khả tách tuyến tính, ta sẽ có những mặt phẳng phân tách những điểm $X_i$ thành 2 lớp riêng biệt. Ta muốn tìm mặt siêu phẳng có thể cực đại độ lớn lề giữa những điểm Y=1 và những điểm Y= -1. Biểu thức thể hiện siêu mặt phẳng:

$$W . X + b = 0$$

Trong đó, W là vector pháp tuyến của mặt phẳng, X là vector đặc trưng, b là hệ số tự do làm mặt phẳng di chuyển tịnh tiến. Hai mặt phẳng lề là hai mặt phẳng hỗ trợ song song với siêu mặt phẳng phân tách và đi qua những điểm nằm gần mặt siêu phẳng phân tách nhất. Những điểm nằm trên hai mặt phẳng hỗ trợ này gọi là các vector hỗ trợ. Hai mặt phẳng lề có dạng:

$$W . X + b = -1$$

$$W . X + b = 1$$

Khoảng cách giữa 2 lề là 2/||W||, với ||W|| là độ dài của vector W.

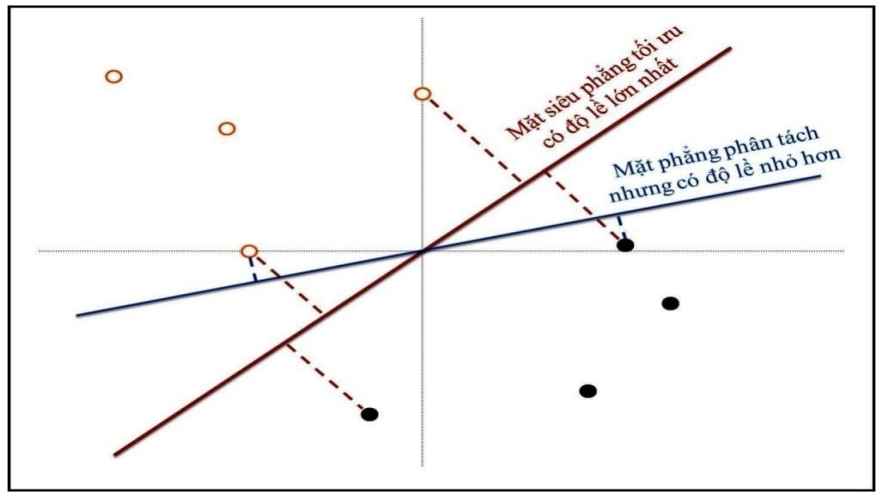

*Hình 1.* Chọn mặt phẳng có thể làm tăng khoảng cách tối đa của biên độ.

Mục tiêu của bài toán là tìm kiếm siêu phẳng phân tách có lề lớn nhất, tức là tìm W và b sao cho cực đại khoảng cách giữa hai mặt phẳng hỗ trợ này, đồng nghĩa với 2/||W|| lớn nhất, tương đương ||W|| là nhỏ nhất. Ngoài ra phải đảm bảo không có điểm dữ liệu nào nằm giữa hai mặt phẳng này, ta thêm các ràng buộc:

$$W \times X_i + b \geq 1, \text{ với } Y_i = +1$$
$$W \times X_i + b \leq -1, \text{ với } Y_i = -1$$

Được viết lại thành:

$$Y(W \times X_i + b) \geq 1, \text{ với } 1 \leq i \leq n$$

Tóm lại cần giải bài toán tối ưu:

$$\begin{cases} \min \|W\| \\ Y_i(W \cdot X_i + b) \geq 1 \end{cases}$$

Việc giải bài toán tối ưu trên là không đơn giản vì ||W|| liên quan đến căn bậc hai nên ta sẽ thay ||W|| với $\frac{1}{2}\|W\|^2$ mà không hề ảnh hưởng đến kết quả, bài toán trở thành bài toán tối ưu quy hoạch toàn phương:

$$\begin{cases} \min \frac{1}{2}\|W\|^2 \\ Y_i(W \cdot X_i + b) \geq 1 \end{cases}$$

## 4. XÂY DỰNG MÔ HÌNH

### 4.1. Sơ đồ dòng dữ liệu

Hình 2 trình bày các bước của quá trình tạo tập đặc trưng cho việc học máy. Tập đặc trưng là một tập hợp các từ thể hiện tâm lý tình cảm có tần số xuất hiện nhiều trong tập huấn luyện. Sau khi có tập đặc trưng thì chúng tôi sẽ kết hợp tập này với tập huấn luyện để tạo ra tập vectơ đặc trưng. Tiếp theo phương pháp học máy sẽ học từ tập vectơ đặc trưng của tập huấn luyện để tạo ra mô hình phân loại phù hợp như thể hiện ở Hình 3. Sau khi có mô hình phân loại các nhận xét thành một trong 3 loại là positive (tích cực), negative (tiêu cực) và neutral (trung lập), hệ thống sẽ tạo tập vectơ đặc trưng cho tập kiểm tra. Tập đặc trưng này sẽ được phân loại bởi mô hình phân loại được xây dựng từ tập kiểm thử. Hình 4 trình bày thứ tự kiểm tra một nhận xét là tích cực, tiêu cực hay trung tính.

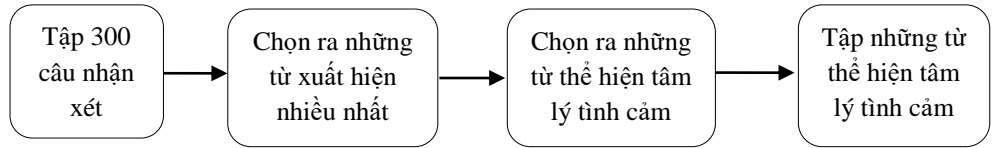

*Hình 2. Sơ đồ dòng dữ liệu cho quá trình tạo tập đặc trưng.*

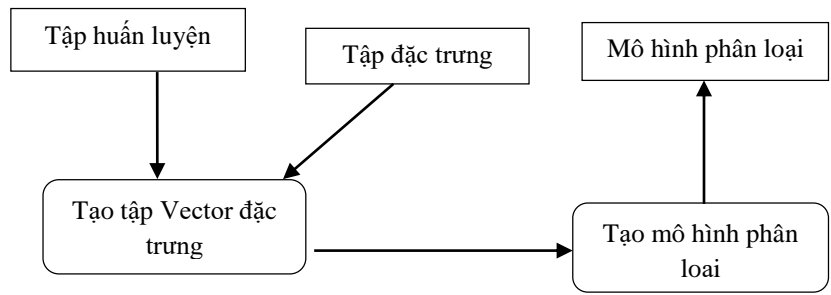

*Hình 3. Sơ đồ dòng dữ liệu cho quá trình tạo mô hình phân loại.*

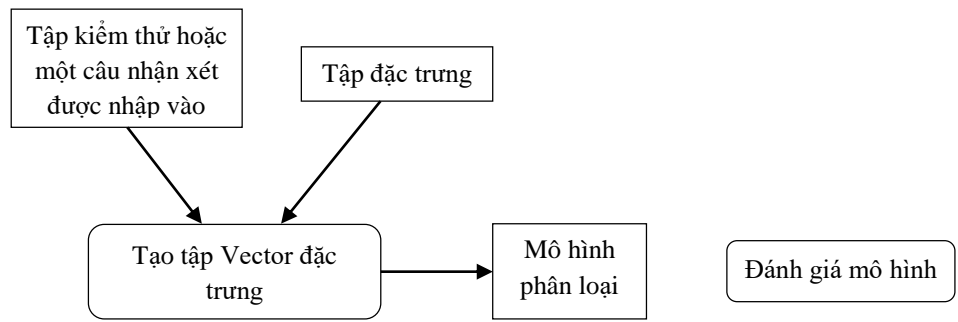

*Hình 4. Sơ đồ dòng dữ liệu cho quá trình đánh giá mô hình phân loại.*

## 4.2. Tập huấn luyện và tập kiểm thử

Chúng tôi thu thập ngẫu nhiên 700 ý kiến đánh giá thuộc 4 loại sản phẩm: giày (shoes), túi xách (handbags), nhẫn (rings), đầm (dresses) trên trang web thương mại điện tử nổi tiếng là Amazon.com. Nội dung tập huấn luyện và tập kiểu thử gồm 6 cột: STT, loại sản phẩm, chủ đề, nhận xét, đánh giá của người và đánh giá của máy. Tiếp theo với mỗi ý kiến, chúng tôi sẽ gán nhãn là tích cực (positive), tiêu cực (negative) hoặc trung lập (neutral) bằng tay và đưa nhãn vào cột đánh giá của người trong tập dữ liệu. Sau đó, chúng tôi chia ngẫu nhiên tập dữ liệu này thành 300 ý kiến cho tập huấn luyện, 400 ý kiến tập kiểm thử.

## 4.3. Mô hình tạo tập đặc trưng và vector chuẩn (FastVector)

Chúng tôi chọn những từ đặc trưng có ảnh hưởng nhất đến việc phân loại và lưu chúng vào file. Các từ đặc trưng này có thể là tính từ (tính chất của sản phẩm), động từ (cảm xúc của người viết), hoặc là danh từ (sự khen ngợi). Các từ đặc trưng này được tạo ra từ 300 câu nhận xét mẫu trên trang amazon.com gồm 4 loại sản phẩm của giới nữ: Đầm (Dresses), Túi xách (Handbags), Nhẫn (Rings), Giày (Shoes). Trong 300 câu nhận xét này ta sẽ tìm ra những từ hay xuất hiện nhất trong các đánh giá. Và từ những từ xuất hiện nhiều nhất ta lại chọn ra những từ thể hiện tâm lý tình cảm có sức ảnh hưởng lớn nhất, có khả năng quyết định nhận xét đó là tích cực, tiêu cực, hay trung lập.

Vector chuẩn (FastVector) có 1 dòng, n cột gọi là n phần tử. Tên của n-1 phần tử đầu tiên là các giá trị (những từ thể hiện cảm nhận của người nhận xét) được đọc từ tập đặc trưng. Và những phần tử này có kiểu dữ liệu là numeric (dạng số), thể hiện số lần suất hiện của phần tử trong nhận xét. Phần tử cuối cùng (thứ n) mang tên Result, được dùng để lưu lại đánh giá của người dùng. Và phần tử cuối cùng này có kiểu dữ liệu là nominal (dạng tập hợp) gồm 3 thuộc tính là positive (tích cực), negative (tiêu cực), và neutral (trung lập). Mục đích để Vector chuẩn này là để dùng làm đầu vào cho việc đánh giá, phân loại trong mô hình học máy.

### 4.4. Mô hình tính trọng số TF-IDF cho từng đặc trưng

Mục đích của việc xác định trọng số cho từng đặc trưng là để tạo ra các vector đặc trưng, nhằm phục vụ cho việc đánh giá trên mô hình học máy. Những vector đặc trưng này sẽ có định dạng giống vector chuẩn. Mỗi vector đặc trưng sẽ biểu diễn cho một đánh giá.

Có 2 phương pháp để tính trọng số:

- Dựa vào việc đặc trưng này có xuất hiện trong đánh giá hay không.

  Giả sử ta có một tập có chứa m đặc trưng $\{t_1, t_2, t_3, t_4, \ldots, t_m\}$, với $d_j$ là câu nhận xét đánh giá thứ j. Gọi $W_{ij}$ là trọng số của đặc trưng thứ i trong đánh giá thứ j. Ta có:

  $$W_{ij} = 0, \text{ nếu } t_i \text{ không thuộc } d_j$$
  $$W_{ij} = 1, \text{ nếu } t_i \text{ thuộc } d_j$$

  Tuy nhiên, cách xác định trọng số này chưa hiệu quả vì độ chính xác không cao nên chúng tôi không chọn phương pháp này.

- Dựa vào tần số xuất hiện của từng đặc trưng trong một đánh giá.

  Số lần xuất hiện của một đặc trưng trong một đánh giá càng lớn và ít xuất hiện trong các câu đánh giá khác thì từ đặc trưng đó có khả năng quyết định kết quả phân loại của đánh giá đó. Theo công thức tính trọng số TF-IDF cho từng đặc trưng trong một câu đánh giá như sau:

  $$tf(t, d) = \frac{f(t,d)}{\max\{f(w, d): w \in d\}}$$

  Trong công thức tính TF thì tf(t, d) là chỉ số tf của đặc trưng t trong đánh giá d. f(t, d) là tần số xuất hiện của đặc trưng t trong câu đánh giá d. max{f(w, d):w $\in$ d} là tần số xuất hiện lớn nhất của một đặc trưng bất kì chứa trong câu đánh giá d. Trong đó d là đánh giá trong tập huấn luyện hay tập kiểm thử tùy vào tập đang xét là tập nào.

  $$idf(t, D) = \log \frac{D}{\{d \in D: t \in d\}+1}$$

Trong công thức IDF thì idf(t, D) là chỉ số idf của đặc trưng t trong tập dữ liệu chứa D đánh giá. D là số câu đánh giá trong tập huấn luyện. {d∈ D: t ∈ d} là số đánh giá trong tổng số D đánh giá có chứa đặc trưng t.

$$W_{(t,d)} = TF \times IDF$$

Trong công thức tính TF-IDF, W(t.d) là trọng số của đặc trưng t trong đánh giá d. Và nó được tính bằng tích của chỉ số TF với chỉ số IDF. Những đặc trưng có chỉ số TF-IDF càng cao thì xuất hiện càng nhiều trong đánh giá và ngược lại những đặc trưng có chỉ số TF-IDF càng thấp thì xuất hiện càng ít trong đánh giá đang xét. Vì vậy mà cách tính này có độ chính xác cao, nên chúng tôi đã sử dụng phương pháp này cho việc tính trọng số.

### 4.5. Mô hình tạo ra 1 kiểu dữ liệu Instances từ tập huấn luyện và tập kiểm thử hay từ một nhận xét được nhập vào bởi người dùng

Instance là một vector chứa định dạng của FastVector được dùng để đại diện cho một đánh giá có n phần tử với n-1 phần tử đầu tiên chứa chỉ số TF-IDF của từng đặctrưng, phần tử thứ n chứa nhãn lớp positive (tích cực), negative (tiêu cực), neutral (trung lập) do người đánh giá. Instances là một tập hợp chứa một hoặc nhiều instance.

Tùy theo người dùng chọn chức năng đánh giá 1 câu nhận xét được nhập vào ô text box hay chức năng đánh giá trên một tập kiểm thử mà ta tạo ra các Instances khác nhau.

- Tạo instances từ tập huấn luyện hay tập kiểm thử: chúng tôi sẽ tính chỉ số TF-IDF của từng đặc trưng trong từng câu nhận xét, sau đó đưa các chỉ TF-IDF này vô n-1 phần tử đầu của một Instance. Phần tử cuối cùng của Instance chúng tôi sẽ gán nhãn lớp (do người đánh giá) vô. Với tập huấn luyện thì chúng tôi có 300 câu đánh giá nên tương ứng tạo ra được 300 Instance, với tập kiểm thử thì chúng tôi có 400 câu đánh giá nên tương ứng tạo ra được 400 Instance. Tiếp theo đó chúng tôi đưa các Instance này vào một Instances để sử dụng làm đầu vào cho mô hình đánh giá học máy.

- Tạo instances từ một câu nhận xét đánh giá: từ 1 câu nhận xét đánh giá được nhập vào text box trên màn hình giao diện chính ta sẽ tính chỉ số TF-IDF của từng đặc trưng. Sau đó, đưa các chỉ số TF-IDF vào n-1 phần tử đầu tiên trong Instance. Phần tử cuối cùng của Instance sẽ là nhãn lớp positive (tích cực), negative (tiêu cực), hay neutral (trung lập). Sau đó đưa Instance này vào Instances làm đầu vào cho mô hình đánh giá bằng phương pháp học máy.

### 4.6. Mô hình đánh giá kết quả dựa trên ma trận ConfusionMatrix

ConfusionMatrix là một tra trận có n hàng n cột. Trong đó n là số nhãn lớp. Ma trận này cho biết số câu được đánh giá và kết quả đánh giá của những câu này.

VD: matrận ConfusionMatrix 3 hàng 3 cột.

$$
\begin{array}{ccc}
4 & 0 & 0 \\
0 & 0 & 3 \\
0 & 1 & 0
\end{array}
$$

Trong đó, tổng các chỉ số trong hàng 1 là số câu positive (tích cực) do người đánh giá. Tổng các chỉ số trong hàng 2 là số câu negative (tiêu cực) do người đánh giá. Tổng các chỉ số trong hàng 3 là số câu neutral (trung lập) do người đánh giá. Tổng các chỉ số trong cột 1 là số câu positive (tích cực) do máy đánh giá. Tổng các chỉ số trong cột 2 là số câu negative (tiêu cực) do máy đánh giá. Tổng các chỉ số trong cột 3 là số câu neutral (trung lập) do máy đánh giá.

Dựa theo ví dụ trên, ở hàng 1 cột 1, chỉ số 4 cho biết rằng người đánh giá 4 câu positive (tích cực) và máy cũng đánh giá 4 câu đó là positive (tích cực) -> máy và người đánh giá giống nhau -> chính xác. Trong hàng 2 cột 3, chỉ số 3 cho biết rằng có 3 câu người đánh giá là negative (tiêu cực) nhưng máy lại đánh giá 3 câu này là neutral (trung lập) -> máy và người đánh giá không giống nhau -> không chính xác. Tương tự trong hàng 3 cột 2, chỉ số 1 cho biết rằng có 1 câu người đánh giá là neutral (trung lập) nhưng máy đánh giá câu này là negative (tiêu cực) -> đánh giá giữa máy và người không giống nhau -> không chính xác.

Như vậy, dựa vào ma trận ConfusionMatrix ta sẽ xác định được kết quả đánh giá của 1 câu nhận xét được nhập vào hay nhiều câu nhận xét trong tập kiểm thử.

### 4.7. Đánh giá mô hình học máy trên tập kiểm thử và một câu nhận xét được người dùng nhập vào

Trước tiên, chúng tôi sẽ khởi tạo mô hình phân loại riêng SMO cho phương pháp máy vector hỗ trợ (SVM). Sau đó, ta sẽ xây dựng việc phân loại dựa trên tập huấn luyện có sẵn. Tiếp sau đó, ta tạo ra mô hình đánh giá dựa trên tập huấn luyện.

Cuối cùng là bước đánh giá mô hình, ta sẽ đánh giá tập kiểm thử dựa trên mô hình phân loại đã tạo ở trên. Rồi xác định kết quả đánh giá dựa vào ma trận ConfusionMatrix đối với 1 Instance (hay 1 câu nhận xét được người dùng nhập vào) hoặc đối với với từng Instance (từng câu trong tập kiểm thử).

Riêng với việc đánh giá trên tập kiểm thử, ta sẽ kết hợp với việc xác định xem có tổng cộng tất cả bao nhiêu câu positive (tích cực), negative (tiêu cực), neutral (trung lập) từ kết quả đánh giá để đưa ra màn hình giao diện chính cho người dùng xem và sử dụng để tính các độ đánh giá.

### 4.8. Mô hình tính độ đánh giá

Ta có TP là số câu người đánh giá và máy đánh giá là giống nhau, FP là số câu người đánh giá và máy đánh giá là không giống nhau, Precision là độ chính xác, Recall là độ đầy đủ, F-measure là độ đầy đủ điều hòa.

Bảng 1 cho biết số câu người và máy cùng đánh giá là tích cực, tiêu cực và trung lập, số câu máy đánh giá là tích cực nhưng người đánh giá là tiêu cực hoặc trung lập, số câu máy đánh giá là tiêu cực nhưng người đánh giá là tích cực hoặc trung lập, số câu máy đánh giá là trung lập nhưng người đánh giá là tích cực hoặc tiêu cực.

TP có 3 loại là:

- TPositive (số câu máy và người đều đánh giá là positive).
- TNegative (số câu máy và người đều đánh giá là negative).
- TNeutral (số câu máy và người đều đánh giá là neutral).

*Bảng 1.* Bảng thể hiện kết quả phân lớp đúng (người đánh giá) và kết quả phân lớp đạt được (máy đánh giá).

| | | Kết quả phân lớp đúng (Người đánh giá) | | |
|---|---|---|---|---|
| | | Positive | Negative | neutral |
| Kết quả phân lớp đạt được (Máy đánh giá) | Positive | TPositive (TP) | FPosNeg (FP) | FPosNeu (FP) |
| | Negative | FNegPos (FP) | TNegative (TP) | FNegNeu (FP) |
| | Neutral | FNeuPos (FP) | FNeuNeg (FP) | TNeutral (TP) |

FP có 6 loại là:

- FPosNeg (số câu máy đánh giá là positive nhưng người đánh giá là negative).
- FPosNeu (số câu máy đánh giá là positive nhưng người đánh giá là neutral).
- FNegPos (số câu máy đánh giá là negative nhưng người đánh giá là positive).
- FNegNeu (số câu máy đánh giá là negative nhưng người đánh giá là neutral).
- FNeuPos (số câu máy đánh giá là neutral nhưng người đánh giá là positive).
- FNeuNeg (số câu máy đánh giá là neutral nhưng người đánh giá là negative).

Precision có 3 chỉ số là:

- PPos (độ chính xác của việc phân loại 1 câu là positive).
- PNeg (độ chính xác của việc phân loại 1 câu là negative).
- PNeu (độ chính xác của việc phân loại 1 câu là neutral).

Recall có 3 chỉ số là:

- RPos (độ đầy đủ của việc phân loại 1 câu là positive).
- RNeg (độ đầy đủ của việc phân loại 1 câu là negative).
- RNeu (độ đầy đủ của việc phân loại 1 câu là neutral).

F-measure có 3 chỉ số là:

- FmPos (độ đầy đủ điều hòa của việc phân loại 1 câu là positive).
- FmNeg (độ đầy đủ điều hòa của việc phân loại 1 câu là negative).
- FmNeu (độ đầy đủ điều hòa của việc phân loại 1 câu là neutral).

Công thức tính độ đánh giá:

- Công thức tính Precision:

$$Precision = \frac{\text{số câu máy đánh giá đúng thuộc lớp i}}{}$$

  - PPos = TPos / (TPos + FPosNeg + FPosNeu)
  - PNeg = TNeg / (TNeg + FNegPos + FNegNeu)
  - PNeu = TNeu / (TNeu + FNeuPos + FNeuNeg)

- Công thức tính Recall:

$$Recall = \frac{\text{số câu máy đánh giá đúng thuộc lớp i}}{}$$

- RPos = TPos / (TPos + FNegPos + FNeuPos)
- RNeg = TNeg / (TNeg + FPosNeg + FNeuNeg)
- RNeu = TNeu / (TNeu + FPosNeu + FNegNeu)
- Công thức tính F-measure:

$$\text{F-measure} = \frac{2 \times \text{Precision} \times \text{Recall}}{\text{Precision} + \text{Recall}}$$

- FmPos = 2 * PPos * RPos / (PPos + RPos)
- FmNeg = 2 * PNeg * RNeg / (PNeg + RNeg)
- FmNeu = 2 * PNeu * RNeu / (PNeu + RNeu)

### 4.9. Kết quả thực nghiệm trên tập kiểm thử

Bảng 2 trình bày kết quả đánh giá, phân loại cho biết số câu người và máy cùng đánh giá là tích cực, tiêu cực và trung lập, số câu máy đánh giá là tích cực nhưng người đánh giá là tiêu cực hoặc trung lập, số câu máy đánh giá là tiêu cực nhưng người đánh giá là tích cực hoặc trung lập, số câu máy đánh giá là trung lập nhưng người đánh giá là tích cực hoặc tiêu cực. Ngoài ra, còn cho biết độ đo Precision tích cực, tiêu cực, trung lập, độ đo Recall tích cực, tiêu cực, trung lập, và độ đo F-measure tích cực, tiêu cực, trung lập.

*Bảng 2.* Bảng kết quả đánh giá hiệu suất mô hình máy vector hỗ trợ trên tập kiểm thử.

| | | Người đánh giá | | | Precision | Recall | F-measure |
|---|---|---|---|---|---|---|---|
| | | Positive | negative | neutral | | | |
| Máy đánh giá | positive | 133 | 1 | 15 | 0.893 | 0.95 | 0.921 |
| | negative | 0 | 102 | 3 | 0.971 | 0.785 | 0.868 |
| | neutral | 7 | 27 | 112 | 0.767 | 0.862 | 0.812 |

Người đánh giá:  positive: 140 câu   negative: 130 câu   neutral: 130 câu

Máy đánh giá:  positive: 149 câu   negative : 105 câu   neutral: 146 câu

## 5. KẾT LUẬN

Trong công trình, chúng tôi đã phân tích tầm quan trọng của nhận xét trong các lĩnh vực khác nhau, thành phần cấu tạo của một đánh giá và so sánh các điểm giống, khác nhau giữa ý kiến nhận xét tiếng Việt với ý kiến nhận xét tiếng Anh. Công trình tập trung vào mục đích phân tích ý kiến nhận xét tiếng Anh là positive (tích cực), negative (tiêu cực) hay neutral (trung lập) dựa vào phương pháp học máy. Vì vậy chúng tôi đã giới thiệu lý thuyết học máy và phương pháp máy vector hỗ trợ (SVM).

Tất cả các ý kiến tiếng Anh dùng để làm tập huấn luyện và tập kiểm thử trong công trình được chúng tôi thu thập từ trang web bán hàng trực tuyến nổi tiếng là www.amazon.com

(Amazon). Chúng tôi đã đề xuất mô hình xây dựng, nêu các bước thực hiện chi tiết, hiện thực mô hình và tiến hành thực nghiệm đánh giá kết quả từ chương trình ứng dụng viết trên nền web Java, trong đó có tích hợp một số thư viện của phần mềm mã nguồn mở weka. Kết quả đánh giá hiệu suất đạt được đối với positive là Precision 89,3 %, Recall 95,0 %, F-measure 92,1 %; đối với negative là Precision 97,1 %, Recall 78,5 %, F-measure 86,8 %; đối với neutral là Precision 76,7 %, Recall 86,2 %, F-measure 81,2 %.

Công trình còn một số hạn chế, chúng tôi đề xuất hướng phát triển như sau:

- Mở rộng tập huấn luyện bằng cách tăng thêm các ý kiến nhận xét và thu thập thêm những ý kiến của nhiều loại sản phẩm khác nhau để làm đa dạng hóa việc phân loại tâm lý người dùng, và mở rộng việc phân loại ở nhiều sản phẩm hơn.

- Cải thiện tập đặc trưng, mở rộng tập đặc trưng ra với một khối từ vựng mở rộng chỉ có ở một hoặc một vài sản phẩm riêng biệt.

- Công trình có thể xử lý được các dạng câu khác nhau như với câu phủ định, câu so sánh và câu điều kiện.

## TÀI LIỆU THAM KHẢO